\documentclass[11pt]{article} 
\usepackage{rldmsubmit,palatino}
\usepackage{graphicx}
\graphicspath{{./figures/}}

\usepackage[T1]{fontenc}
\usepackage{authblk}

\usepackage[square,numbers]{natbib}
\usepackage[hidelinks]{hyperref}
\usepackage{amsmath,amssymb}
\usepackage{bbm}
\usepackage{subcaption}
\usepackage{algorithm}
\usepackage[noend]{algpseudocode}

\title{All You Need Is Supervised Learning: From Imitation Learning to Meta-RL With Upside Down RL}

\author[$\;\;$,1,2]{Kai Arulkumaran\thanks{Correspondence to \texttt{kai\_arulkumaran@araya.org}. Work partially done while author was at NNAISENSE.}}
\author[$\,$3,4,5]{Dylan R. Ashley}
\author[$\,$3,4,5,6,7]{J{\"u}rgen Schmidhuber}
\author[$\,$7]{Rupesh K. Srivastava}
\affil[1$\,$]{ARAYA, Inc., Tokyo, Japan}
\affil[2$\,$]{Imperial College London, London, UK}
\affil[3$\,$]{The Swiss AI Lab IDSIA, Lugano, Switzerland}
\affil[4$\,$]{Universit{\`{a}} della Svizzera Italiana (USI), Lugano, Switzerland}
\affil[5$\,$]{Scuola Universitaria Professionale della Svizzera Italiana (SUPSI), Lugano, Switzerland}
\affil[6$\,$]{King Abdullah University of Science and Technology (KAUST), Thuwal, Saudi Arabia}
\affil[7$\,$]{NNAISENSE, Lugano, Switzerland}

\begin{document}

\maketitle

\begin{abstract}
Upside down reinforcement learning (UDRL) flips the conventional use of the return in the objective function in RL upside down, by taking returns as input and predicting actions. UDRL is based purely on supervised learning, and bypasses some prominent issues in RL: bootstrapping, off-policy corrections, and discount factors. While previous work with UDRL demonstrated it in a traditional online RL setting, here we show that this single algorithm can also work in the imitation learning and offline RL settings, be extended to the goal-conditioned RL setting, and even the meta-RL setting. With a general agent architecture, a single UDRL agent can learn across all paradigms.

\end{abstract}


\acknowledgements{This work was partially supported by the European Research Council (ERC, Advanced Grant Number 742870).}

\startmain 

\section{Introduction}
\vspace{-0.6em}

Over time, the field of AI has exhibited significant convergence, with the increasing popularity of artificial neural networks (NNs) trained on large amounts of data resulting in improved performance across many benchmarks. While many earlier successes were based on supervised learning (SL), NNs have also revolutionised reinforcement learning (RL).

We begin by noting two further significant convergences in recent years. The first pertains to NN architectures. While convolutional NNs (CNNs) and recurrent NNs (RNNs) have for decades been applied to images and text, respectively, a new general-purpose architecture has emerged \cite{jaegle2021perceiver}. Originally applied to text, the Transformer architecture \cite{vaswani2017attention} has been successfully extended to images \cite{dosovitskiy2021image}, as well as other structured input/outputs \cite{jaegle2021perceiver}. The advantage of such models is the ability to operate over \emph{sets}, allowing a single, modular model to reuse knowledge over a dynamic set of inputs.

The second development is the rise of self-supervised learning (SSL) across different domains \cite{weng2019selfsup}: the use of ``pretext tasks'', constructing labels from unlabelled data in order to imitate SL training, has resulted in models that rival those trained with SL. SSL benefits from rich ``supervisory signals'', with models learning to \emph{predict} transformations, $f$, of their inputs, $x$; loosely speaking, $p(f(x)|x)$, versus SL's $p(y|x)$, where $y$ is a label. In a single domain, a variety of tasks can enable a model to gain complementary knowledge over different facets of the world.

We propose uniting these directions to create general learning agents, with the position that RL can itself be framed as an SL problem. This is not a novel proposition \cite{peters2007reinforcement,schmidhuber2019reinforcement,srivastava2019training,kumar2019reward,ghosh2021learning,chen2021decision,janner2021reinforcement,emmons2021rvs,furuta2022generalized}, but in contrast to prior works, we provide a general framework that includes online RL, goal-conditioned RL (GCRL) \cite{schmidhuber1990learning}, imitation learning (IL) \cite{pomerleau1988alvinn}, offline RL \cite{ernst2005tree}, and meta-RL \cite{schmidhuber1994learning}, as well as other paradigms contained within partially observed Markov decision processes (POMDPs) \cite{ni2021recurrent}. We build upon the proposal of upside down RL (UDRL) by Schmidhuber \cite{schmidhuber2019reinforcement}, the implementation of Srivastava et al. \cite{srivastava2019training}, and sequence modelling via Decision Transformers \cite{chen2021decision,janner2021reinforcement}. We examine current implementations, discuss a generalisation of the framework, and then demonstrate a \emph{single algorithm and architecture} on a standard control problem.

\section{Upside Down RL}
\vspace{-0.6em}

The core of UDRL is the policy, $\pi$, conditioned on ``commands'', $c$ \cite{schmidhuber2019reinforcement}. Given a dataset $D$ of trajectories (states, $s$, actions, $a$, and rewards, $r$), the policy, parameterised by $\theta$, is trained using an SL loss, $\mathcal{L}$, to map states and commands to actions:
\begin{align}
  \arg\min_\theta \mathbb{E}_{s,a,r \sim D}[\mathcal{L}(a, \pi(a|s, c; \theta))].
\end{align}
In the initial implementation \cite{srivastava2019training}, $c = [d^H, d^R]$, where $d^H$ is the desired (future) time horizon, $t_2 - t_1$, and $d^R$ is the (time-bounded) return-to-go, $\sum_{t=t_1}^{t_2}r_t$. In a similar fashion to hindsight experience replay\footnote{But without specific environment settings, e.g., symbolic state spaces or a distance-based goal-conditional reward functions.} \cite{kaelbling1993learning,andrychowicz2017hindsight}, the agent can be trained on data sampled from $D$, calculating $d^H$ and $d^R$ directly from the data. Trained thus, the agent can then achieve a desired return by sampling actions from its stochastic command-conditioned policy. Unlike typical RL agents, UDRL agents are capable of achieving low returns, medium returns, or high returns, based on the choice of $d^R$ \cite{srivastava2019training}.

There are two key elements to the efficacy of UDRL---the first being the dataset (and its use) \cite{riedmiller2022collect}. In the online RL setting, a UDRL agent can learn to perform reward maximisation in a manner akin to expectation maximisation: collect data using the policy, then train on the most rewarding data. This can be achieved by weighting the data \cite{peters2007reinforcement}, or controlling data storage/sampling \cite{srivastava2019training}. Given an existing dataset of demonstrations, the agent can be trained in the offline RL setting \cite{kumar2019reward,janner2021reinforcement,chen2021decision}. $D$ can be used for auxiliary tasks to improve learning/generalisation, such as via learning a world model \cite{janner2021reinforcement}.

The second key element is the use of commands. $c$ can be any computable predicate that is consistent with the data \cite{schmidhuber2019reinforcement}. In the initial implementation, the agent is trained to map observed actions $a_t$ to the corresponding states $s_t$ and $[d^H, d^R]$, where the latter is calculated from $t$ to the terminal timestep $T$, allowing the agent to perform (undiscounted) credit assignment across the entire episode. During exploration in the environment (which generates more data for the agent), $d^H$ is set to the mean of the most rewarding episodes in $D$, and $d^R$ is sampled uniformly from values between the mean of the most rewarding episodes' returns, and the mean plus one standard deviation (encouraging optimism). After every environment interaction, $d^H$ is decremented by 1 and $d^R$ is decremented by $r$. If $c$ is simplified to only contain the desired return, we recover reward-conditioned policies \cite{kumar2019reward}, and if $c = \emptyset$, we recover behavioural cloning (BC) \cite{pomerleau1988alvinn}, which is the simplest IL algorithm. Conversely, the trivial augmentation of $c$ with a goal vector $g$ extends UDRL to the GCRL setting \cite{schmidhuber2019reinforcement,ghosh2021learning,janner2021reinforcement}, with zero-shot generalisation enabled via appropriate goal spaces (e.g., language \cite{jiang2019language}).

As the agent is trained using SL on observed data, UDRL bypasses several commmon issues in RL: bootstrapping (temporal-difference updates), off-policy corrections, and discount factors. This allows us to more easily focus on standard points in ML: the agent's ability to generalise to novel states and commands is based on the data available, the class of policies, and the optimisation process \cite{emmons2021rvs}. Srivastava et al.\,\cite{srivastava2019training} trained fully-connected-/CNN-based UDRL agents using stochastic gradient descent on the cross-entropy loss, but replacing the architecture (e.g., with Transformers \cite{janner2021reinforcement,chen2021decision}) or the optimisation (e.g., with evolutionary algorithms) are valid alternatives. While the move to sequence modelling allows the agent to benefit from greater contexts \cite{janner2021reinforcement,chen2021decision}, in general, a \emph{stateful} agent is needed in order to deal with POMDPs, as well as more general computable predicates \cite{schmidhuber2019reinforcement}.

\section{Generalised Upside Down RL in POMDPs}
\vspace{-0.6em}

While many realistic problems cannot be captured by MDPs, they can be reasonably modelled by POMDPs, in which the agent only receives a partial observation $o$ of the true state $s$. A principled approach to solving POMDPs is to keep a ``belief'' over the state, which can be achieved implicitly by training an RNN agent, which updates a hidden state vector $h$ \cite{wierstra2007solving}. POMDPs encompass many other problems, such as hidden parameter MDPs \cite{doshi2016hidden}, which consider related tasks, and even the general meta-RL setting (where $h$ is typically augmented with previous action, reward and terminal indicator, $\mathbbm{1}_\text{terminal}$ \cite{duan2016rl,wang2017learning}). POMDPs can also be related to generalisation in RL and the robust RL \cite{morimoto2000robust} setting, which considers worst-case performance. While various specialised algorithms exist for these different problem settings, recent results from Ni et al. \cite{ni2021recurrent} have shown that simple recurrent model-free RL agents are can perform well across the board. Motivated by this (and further related arguments by Schmidhuber \cite{schmidhuber2019reinforcement}), we proceed by treating every environment as a POMDP, in which an agent attempts to learn using a general algorithmic framework.

A final ingredient enables a \emph{single agent} to deal with all RL problem settings (plus IL) with \emph{one model} --- an architecture that can deal with arbitrary structured inputs and outputs (e.g., Perceiver IO \cite{jaegle2021perceiver}). Such a model allows $c$ to be dynamic as needed: being null in the pure IL setting (where UDRL reduces to BC), being $[d^H, d^R, g]$ in the GCRL setting, and extending further beyond to incorporate SSL tasks. This alleviates the problem of using unlabelled demonstrations to bootstrap an RL agent, as the agent does not need to model the rewards achieved. Furthermore, the ability to adapt to different observation and action spaces allows such an agent to use third-person data for representation learning.\footnote{Correspondences between the agent and third party's observation/action spaces would be needed for true third-person IL \cite{stadie2017third}. To maintain such flexibility, the previous action and reward can be included in a dynamic $c$ structure, instead of within $h$.}

Given this, we present a generalised algorithm for UDRL (Algorithm \ref{alg:udrl}). Although nearly all RL problems consider the episodic setting (in which a terminal indicator is given before the environment resets), the following algorithm is applicable in both episodic and non-episodic MDPs, making it capable of continual/lifelong learning. 

\begin{algorithm}
\footnotesize
\caption{Generalised Upside Down RL}\label{alg:udrl}
\begin{algorithmic}
\Require $E$ \Comment{POMDP environment}
\Require $\pi(a|o, c, h)$ \Comment{Command-conditioned recurrent policy}
\Require $D$ \Comment{Experience replay memory; existing data optional unless performing IL/offline RL}
\vspace{0.4em}
\Function{Reset}{$E, \pi, D$}
  \State Reset environment $E$ and $\pi$'s hidden state $h$ \Comment{$h$ also contains the previous action, reward, and terminal indicator}
  \State Get initial observation and goal $(o, g)$ from $E$
  \State Sample $c$ based on $D$ and $(o, g)$ \Comment{Requires a procedure for sampling an initial command $c$ based on observed data}
\EndFunction
\vspace{0.4em}
\State Train $\pi$ on batches from $D$ \Comment{Data can be non-uniformly/adaptively sampled from $D$; auxiliary objectives can be used}
\If{performing IL or offline RL without environment interaction}
  \Return
\EndIf
\vspace{0.4em}
\State \Call{Reset}{$E, \pi, D$}
\While{true}
  \State Act with $a, h \sim \pi(a|o, c, h)$
  \State Observe $(o', r, g', \mathbbm{1}_\text{terminal})$ from environment transition \Comment{$g'$ given for goal-conditioned RL, $\mathbbm{1}_\text{terminal}$ for an episodic MDP}
  \State Update $D$ with $(o, a, r, g, \mathbbm{1}_\text{terminal})$ \Comment{$D$ may prioritise updates/remove old data}
  \State Update $h$ (to contain $a$ and $r$) and $c$ \Comment{Requires a procedure for updating $c$, e.g., include $g$, decrement $d^H$}
  \State Train $\pi$ on batches from $D$
  \If{$\mathbbm{1}_\text{terminal}$}
    \Call{Reset}{$E, \pi, D$}
  \EndIf
\EndWhile
\end{algorithmic}
\end{algorithm}

\section{Experiments}
\vspace{-0.6em}

For our experiments, we use the classic CartPole control problem \cite{barto1983neuronlike}, where the agent receives a +1 reward for every timestep the pole is balanced, with a time limit of $500$ timesteps. We adapt this environment to demonstrate how a single architecture can be used in the following settings: online RL, IL, offline RL, GCRL, and meta-RL. For the IL setting, we train the agent on 5 episodes with the maximum return of $500$, collected from an online RL agent, and disable the reward and desired return inputs. For the offline setting, we train the agent on the worst $1000$ trajectories from the online agent, with an average return of $162 \pm 195$. In the GCRL setting, at the beginning of each episode the agent is given an $x$ goal position uniformly sampled from $[-1, 1]$, with the reward function set to $e^{-|x - g|}$. In the meta-RL setting, several environment parameters are randomly sampled at the beginning of each episode \cite{li2018learning}. In the GCRL and meta-RL settings, during testing the agent is evaluated on a cross product of a uniform spread over goals/environment parameters.

As the observation and action spaces are constant, we use a simple base policy that consists of a linear layer, a sigmoid-gated linear layer, an LSTM \cite{hochreiter1997long}, and a final linear layer to predict the logits of a categorical distribution. Each command ($d^H$ and $d^R$), the goal, previous action, reward, and terminal indicator, are all embedded and concatenated with a learnable encoding vector, before being processed by a Transformer encoder layer \cite{vaswani2017attention}; the resulting vectors are aggregated using the $\max$ function and then used in the gated linear layer before the LSTM. For simplicity, all settings use the same architecture and hyperparameters; the code is available at \url{https://github.com/Kaixhin/GUDRL}.

\begin{figure}
  \centering
  \begin{subfigure}{0.195\textwidth}
    \centering
    \includegraphics[width=\textwidth]{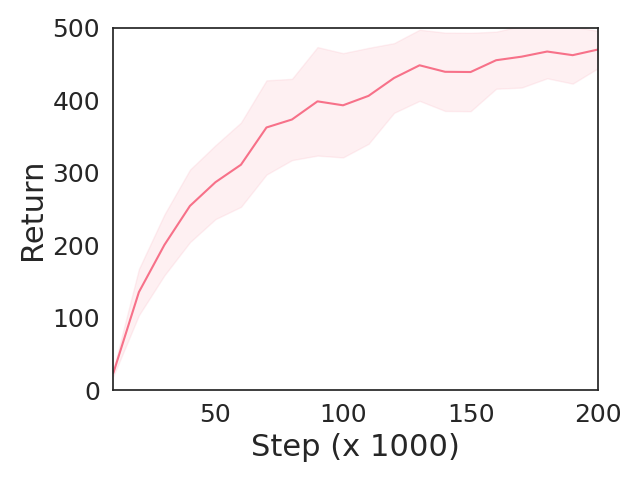}
    \caption{Online RL}
    \label{fig:online}
  \end{subfigure}
  \hfill
  \begin{subfigure}{0.195\textwidth}
    \centering
    \includegraphics[width=\textwidth]{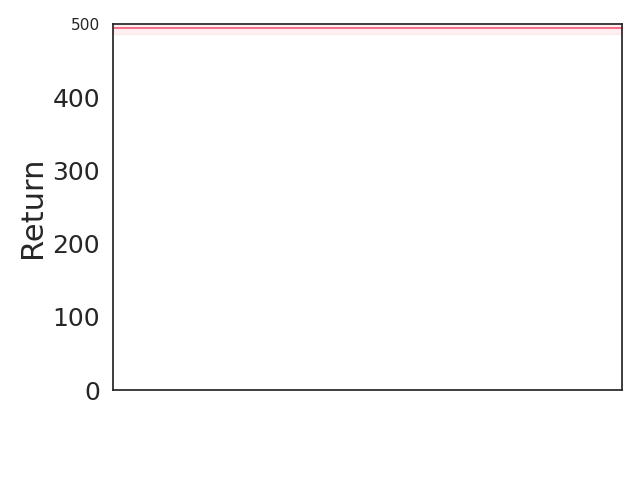}
    \caption{IL}
    \label{fig:imitation}
  \end{subfigure}
  \hfill
  \begin{subfigure}{0.195\textwidth}
    \centering
    \includegraphics[width=\textwidth]{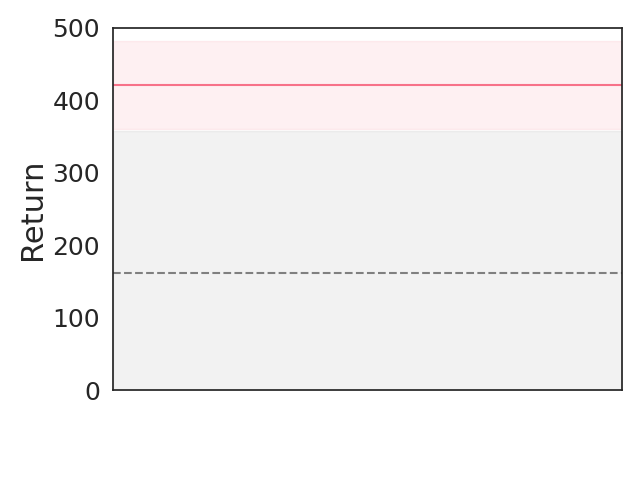}
    \caption{Offline RL}
    \label{fig:offline}
  \end{subfigure}
  \hfill
  \begin{subfigure}{0.195\textwidth}
    \centering
    \includegraphics[width=\textwidth]{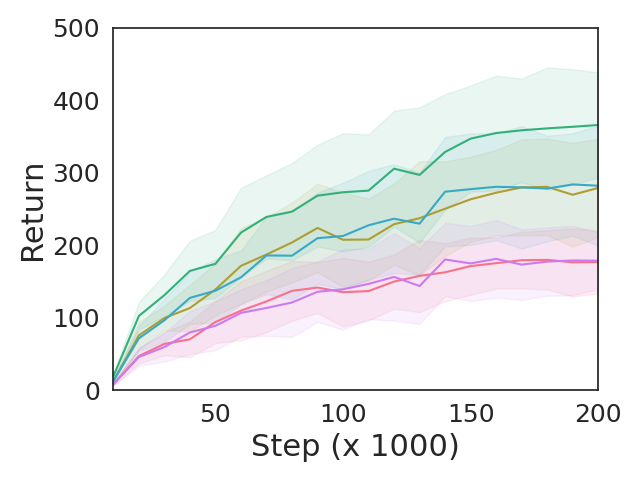}
    \caption{GCRL}
    \label{fig:goal}
  \end{subfigure}
  \hfill
  \begin{subfigure}{0.195\textwidth}
    \centering
    \includegraphics[width=\textwidth]{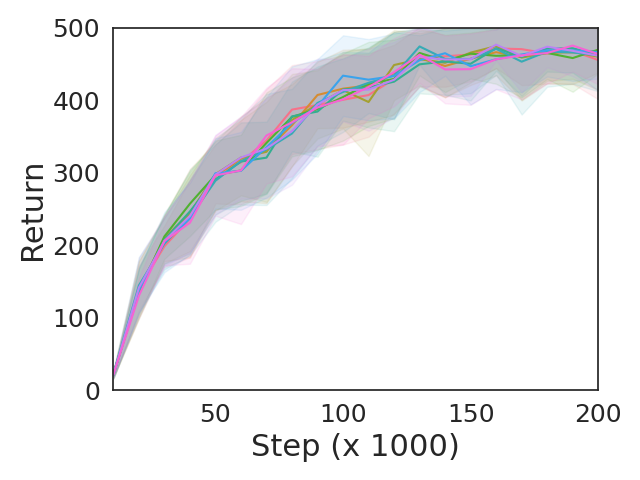}
    \caption{Meta-RL}
    \label{fig:meta}
  \end{subfigure}
  \caption{Generalised UDRL agent trained under the (a) online RL (b) IL (c) offline RL (d) GCRL and (e) meta-RL settings in the CartPole environment. Results averaged over 10 test episodes per evaluation $\times$ 20 random seeds. The dashed line in (c) represents the distribution of returns in $D$. The different colours in (d) and (e) represent different goals/environment parameters, respectively.}
  \label{fig:results}
\end{figure}

As shown in Figure \ref{fig:results}, the generalised UDRL agent is able to learn under all settings. While its performance may not match more specialised RL algorithms, it demonstrates that a single SL objective is sufficient for a variety of sequential decision making problems. Whilst it is possible to tune hyperparameters for individual tasks, a more compelling avenue for improving the performance of such agents is to incorporate further commands/tasks that relate to the environment's structure. This can include learning world models, SSL tasks \cite{weng2019selfsup}, and more general computable predicates \cite{schmidhuber2019reinforcement}.

\section{Discussion}
\vspace{-0.6em}

Given the increased interest in the RL-as-SL paradigm, this work aims to construct a more general purpose agent/learning algorithm, but with more concrete implementation details and links to existing RL concepts than prior work \cite{schmidhuber2019reinforcement}. A major question is whether such an idea can scale? As mentioned previously, optimisation and function approximation are key limiters. Other experiments with tabular representations have yielded UDRL agents that can learn more complex commands \cite{schmidhuber2019reinforcement}; and in further experiments, the act of resetting weights has been useful for more complex agents. With less of the confounding problems of other RL algorithms, UDRL lays bare the problem of continual learning (and proactive interference, in particular) \cite{fedus2020catastrophic,igl2021transient}.

Another discussion point UDRL introduces is the method by which (value) credit assignment can occur. While other agents typically consider the (discounted) episodic return, UDRL agents can incorporate a desired horizon. As such, UDRL might lend itself better to hierarchical RL \cite{gurtler2021hierarchical}. In the current formulation, desired returns and goals are almost interchangeable in the command structure, but a more powerful formulation is to consider these as being variables that can be inferred themselves, i.e., a joint model $(o, a, d^R, g)$. In the same way that inverse models can complement forward models, the joint distribution can interchange $Q$-functions, $Q(d^R|o, a, g)$ and return-conditioned policies, $\pi(a|o, d^R, g)$, utilising whichever is better for the situation (acting, or learning, in a data-dependent manner). Being able to infer the desired goal becomes particularly important when it comes to generalisation in the IL setting \cite{de2019causal}, and could even allow the use of ``suboptimal'' data \cite{emmons2021rvs}. Looking even further forward, with both hierarchy and a more intelligent command selection strategy, UDRL agents could be powerful vessels for implementing open-ended, goal-driven curiosity \cite{oudeyer2007intrinsic,schmidhuber2010formal}.

\bibliography{references}

\begin{thebibliography}{37}
\providecommand{\natexlab}[1]{#1}
\providecommand{\url}[1]{\texttt{#1}}
\expandafter\ifx\csname urlstyle\endcsname\relax
  \providecommand{\doi}[1]{doi: #1}\else
  \providecommand{\doi}{doi: \begingroup \urlstyle{rm}\Url}\fi

\bibitem[Andrychowicz et~al.(2017)Andrychowicz, Wolski, Ray, Schneider, Fong,
  Welinder, McGrew, Tobin, Pieter~Abbeel, and
  Zaremba]{andrychowicz2017hindsight}
M.~Andrychowicz, F.~Wolski, A.~Ray, J.~Schneider, R.~Fong, P.~Welinder,
  B.~McGrew, J.~Tobin, O.~Pieter~Abbeel, and W.~Zaremba.
\newblock {Hindsight Experience Replay}.
\newblock In \emph{NeurIPS}, 2017.

\bibitem[Barto et~al.(1983)Barto, Sutton, and Anderson]{barto1983neuronlike}
A.~G. Barto, R.~S. Sutton, and C.~W. Anderson.
\newblock {Neuronlike Adaptive Elements That Can Solve Difficult Learning
  Control Problems}.
\newblock \emph{IEEE Trans. SMC}, 13\penalty0 (5):\penalty0 834--846, 1983.

\bibitem[Chen et~al.(2021)Chen, Lu, Rajeswaran, Lee, Grover, Laskin, Abbeel,
  Srinivas, and Mordatch]{chen2021decision}
L.~Chen, K.~Lu, A.~Rajeswaran, K.~Lee, A.~Grover, M.~Laskin, P.~Abbeel,
  A.~Srinivas, and I.~Mordatch.
\newblock {Decision Transformer: Reinforcement Learning via Sequence Modeling}.
\newblock \emph{arXiv:2106.01345}, 2021.

\bibitem[De~Haan et~al.(2019)De~Haan, Jayaraman, and Levine]{de2019causal}
P.~De~Haan, D.~Jayaraman, and S.~Levine.
\newblock {Causal Confusion in Imitation Learning}.
\newblock \emph{NeurIPS}, 2019.

\bibitem[Doshi-Velez and Konidaris(2016)]{doshi2016hidden}
F.~Doshi-Velez and G.~Konidaris.
\newblock {Hidden Parameter Markov Decision Processes: A Semiparametric
  Regression Approach for Discovering Latent Task Parametrizations}.
\newblock In \emph{IJCAI}, 2016.

\bibitem[Dosovitskiy et~al.(2021)Dosovitskiy, Beyer, Kolesnikov, Weissenborn,
  Zhai, Unterthiner, Dehghani, Minderer, Heigold, Gelly,
  et~al.]{dosovitskiy2021image}
A.~Dosovitskiy, L.~Beyer, A.~Kolesnikov, D.~Weissenborn, X.~Zhai,
  T.~Unterthiner, M.~Dehghani, M.~Minderer, G.~Heigold, S.~Gelly, et~al.
\newblock {An Image is Worth 16x16 Words: Transformers for Image Recognition at
  Scale}.
\newblock In \emph{ICLR}, 2021.

\bibitem[Duan et~al.(2016)Duan, Schulman, Chen, Bartlett, Sutskever, and
  Abbeel]{duan2016rl}
Y.~Duan, J.~Schulman, X.~Chen, P.~L. Bartlett, I.~Sutskever, and P.~Abbeel.
\newblock {RL$^2$: Fast Reinforcement Learning via Slow Reinforcement
  Learning}.
\newblock \emph{arXiv:1611.02779}, 2016.

\bibitem[Emmons et~al.(2021)Emmons, Eysenbach, Kostrikov, and
  Levine]{emmons2021rvs}
S.~Emmons, B.~Eysenbach, I.~Kostrikov, and S.~Levine.
\newblock {RvS: What is Essential for Offline RL via Supervised Learning?}
\newblock \emph{arXiv:2112.10751}, 2021.

\bibitem[Ernst et~al.(2005)Ernst, Geurts, and Wehenkel]{ernst2005tree}
D.~Ernst, P.~Geurts, and L.~Wehenkel.
\newblock {Tree-based Batch Mode Reinforcement Learning}.
\newblock \emph{JMLR}, 6:\penalty0 503--556, 2005.

\bibitem[Fedus et~al.(2020)Fedus, Ghosh, Martin, Bellemare, Bengio, and
  Larochelle]{fedus2020catastrophic}
W.~Fedus, D.~Ghosh, J.~D. Martin, M.~G. Bellemare, Y.~Bengio, and
  H.~Larochelle.
\newblock {On Catastrophic Interference in Atari 2600 Games}.
\newblock \emph{arXiv:2002.12499}, 2020.

\bibitem[Furuta et~al.(2022)Furuta, Matsuo, and Gu]{furuta2022generalized}
H.~Furuta, Y.~Matsuo, and S.~S. Gu.
\newblock {Generalized Decision Transformer for Offline Hindsight Information
  Matching}.
\newblock In \emph{ICLR}, 2022.

\bibitem[Ghosh et~al.(2021)Ghosh, Gupta, Reddy, Fu, Devin, Eysenbach, and
  Levine]{ghosh2021learning}
D.~Ghosh, A.~Gupta, A.~Reddy, J.~Fu, C.~M. Devin, B.~Eysenbach, and S.~Levine.
\newblock {Learning to Reach Goals via Iterated Supervised Learning}.
\newblock In \emph{ICLR}, 2021.

\bibitem[G{\"u}rtler et~al.(2021)G{\"u}rtler, B{\"u}chler, and
  Martius]{gurtler2021hierarchical}
N.~G{\"u}rtler, D.~B{\"u}chler, and G.~Martius.
\newblock {Hierarchical Reinforcement Learning with Timed Subgoals}.
\newblock In \emph{NeurIPS}, 2021.

\bibitem[Hochreiter and Schmidhuber(1997)]{hochreiter1997long}
S.~Hochreiter and J.~Schmidhuber.
\newblock {Long Short-term Memory}.
\newblock \emph{Neural Comput.}, 9\penalty0 (8):\penalty0 1735--1780, 1997.

\bibitem[Igl et~al.(2021)Igl, Farquhar, Luketina, Boehmer, and
  Whiteson]{igl2021transient}
M.~Igl, G.~Farquhar, J.~Luketina, W.~Boehmer, and S.~Whiteson.
\newblock {Transient Non-stationarity and Generalisation in Deep Reinforcement
  Learning}.
\newblock In \emph{ICLR}, 2021.

\bibitem[Jaegle et~al.(2021)Jaegle, Borgeaud, Alayrac, Doersch, Ionescu, Ding,
  Koppula, Zoran, Brock, Shelhamer, et~al.]{jaegle2021perceiver}
A.~Jaegle, S.~Borgeaud, J.-B. Alayrac, C.~Doersch, C.~Ionescu, D.~Ding,
  S.~Koppula, D.~Zoran, A.~Brock, E.~Shelhamer, et~al.
\newblock {Perceiver IO: A General Architecture for Structured Inputs \&
  Outputs}.
\newblock \emph{arXiv:2107.14795}, 2021.

\bibitem[Janner et~al.(2021)Janner, Li, and Levine]{janner2021reinforcement}
M.~Janner, Q.~Li, and S.~Levine.
\newblock {Offline Reinforcement Learning as One Big Sequence Modeling
  Problem}.
\newblock In \emph{NeurIPS}, 2021.

\bibitem[Jiang et~al.(2019)Jiang, Gu, Murphy, and Finn]{jiang2019language}
Y.~Jiang, S.~S. Gu, K.~P. Murphy, and C.~Finn.
\newblock {Language as an Abstraction for Hierarchical Deep Reinforcement
  Learning}.
\newblock \emph{NeurIPS}, 2019.

\bibitem[Kaelbling(1993)]{kaelbling1993learning}
L.~P. Kaelbling.
\newblock {Learning to Achieve Goals}.
\newblock In \emph{IJCAI}, 1993.

\bibitem[Kumar et~al.(2019)Kumar, Peng, and Levine]{kumar2019reward}
A.~Kumar, X.~B. Peng, and S.~Levine.
\newblock {Reward-conditioned Policies}.
\newblock \emph{arXiv:1912.13465}, 2019.

\bibitem[Li et~al.(2018)Li, Yang, Song, and Hospedales]{li2018learning}
D.~Li, Y.~Yang, Y.-Z. Song, and T.~M. Hospedales.
\newblock {Learning to Generalize: Meta-learning for Domain Generalization}.
\newblock In \emph{AAAI}, 2018.

\bibitem[Morimoto and Doya(2000)]{morimoto2000robust}
J.~Morimoto and K.~Doya.
\newblock {Robust Reinforcement Learning}.
\newblock In \emph{NeurIPS}, 2000.

\bibitem[Ni et~al.(2021)Ni, Eysenbach, and Salakhutdinov]{ni2021recurrent}
T.~Ni, B.~Eysenbach, and R.~Salakhutdinov.
\newblock {Recurrent Model-free RL is a Strong Baseline for Many POMDPs}.
\newblock \emph{arXiv:2110.05038}, 2021.

\bibitem[Oudeyer et~al.(2007)Oudeyer, Kaplan, and Hafner]{oudeyer2007intrinsic}
P.-Y. Oudeyer, F.~Kaplan, and V.~V. Hafner.
\newblock {Intrinsic Motivation Systems for Autonomous Mental Development}.
\newblock \emph{IEEE Trans. Evol. Comput.}, 11\penalty0 (2):\penalty0 265--286,
  2007.

\bibitem[Peters and Schaal(2007)]{peters2007reinforcement}
J.~Peters and S.~Schaal.
\newblock {Reinforcement Learning by Reward-weighted Regression for Operational
  Space Control}.
\newblock In \emph{ICML}, pages 745--750, 2007.

\bibitem[Pomerleau(1988)]{pomerleau1988alvinn}
D.~A. Pomerleau.
\newblock {ALVINN: An Autonomous Land Vehicle in a Neural Network}.
\newblock In \emph{NeurIPS}, 1988.

\bibitem[Riedmiller et~al.(2022)Riedmiller, Springenberg, Hafner, and
  Heess]{riedmiller2022collect}
M.~Riedmiller, J.~T. Springenberg, R.~Hafner, and N.~Heess.
\newblock {Collect \& Infer-A Fresh Look at Data-efficient Reinforcement
  Learning}.
\newblock In \emph{CoRL}, 2022.

\bibitem[Schmidhuber(1990)]{schmidhuber1990learning}
J.~Schmidhuber.
\newblock {Learning Algorithms for Networks with Internal and External
  Feedback}.
\newblock In \emph{Connectionist Models}, pages 52--61. Elsevier, 1990.

\bibitem[Schmidhuber(1994)]{schmidhuber1994learning}
J.~Schmidhuber.
\newblock {On Learning How to Learn Learning Strategies}.
\newblock Technical Report FKI-198-94, Technische Universität München, 1994.

\bibitem[Schmidhuber(2010)]{schmidhuber2010formal}
J.~Schmidhuber.
\newblock {Formal Theory of Creativity, Fun, and Intrinsic Motivation
  (1990--2010)}.
\newblock \emph{IEEE TAMD}, 2\penalty0 (3):\penalty0 230--247, 2010.

\bibitem[Schmidhuber(2019)]{schmidhuber2019reinforcement}
J.~Schmidhuber.
\newblock {Reinforcement Learning Upside Down: Don't Predict Rewards--Just Map
  Them to Actions}.
\newblock \emph{arXiv:1912.02875}, 2019.

\bibitem[Srivastava et~al.(2019)Srivastava, Shyam, Mutz, Ja{\'s}kowski, and
  Schmidhuber]{srivastava2019training}
R.~K. Srivastava, P.~Shyam, F.~Mutz, W.~Ja{\'s}kowski, and J.~Schmidhuber.
\newblock {Training Agents Using Upside-down Reinforcement Learning}.
\newblock In \emph{NeurIPS Deep RL Workshop}, 2019.

\bibitem[Stadie et~al.(2017)Stadie, Abbeel, and Sutskever]{stadie2017third}
B.~C. Stadie, P.~Abbeel, and I.~Sutskever.
\newblock {Third-person Imitation Learning}.
\newblock In \emph{ICLR}, 2017.

\bibitem[Vaswani et~al.(2017)Vaswani, Shazeer, Parmar, Uszkoreit, Jones, Gomez,
  Kaiser, and Polosukhin]{vaswani2017attention}
A.~Vaswani, N.~Shazeer, N.~Parmar, J.~Uszkoreit, L.~Jones, A.~N. Gomez,
  {\L}.~Kaiser, and I.~Polosukhin.
\newblock {Attention Is All You Need}.
\newblock In \emph{NeurIPS}, 2017.

\bibitem[Wang et~al.(2017)Wang, Kurth-Nelson, Soyer, Leibo, Tirumala, Munos,
  Blundell, Kumaran, and Botvinick]{wang2017learning}
J.~X. Wang, Z.~Kurth-Nelson, H.~Soyer, J.~Z. Leibo, D.~Tirumala, R.~Munos,
  C.~Blundell, D.~Kumaran, and M.~M. Botvinick.
\newblock {Learning to Reinforcement Learn}.
\newblock In \emph{CogSci}, 2017.

\bibitem[Weng(2019)]{weng2019selfsup}
L.~Weng.
\newblock {Self-Supervised Representation Learning}.
\newblock \emph{lilianweng.github.io/lil-log}, 2019.

\bibitem[Wierstra et~al.(2007)Wierstra, Foerster, Peters, and
  Schmidhuber]{wierstra2007solving}
D.~Wierstra, A.~Foerster, J.~Peters, and J.~Schmidhuber.
\newblock {Solving Deep Memory POMDPs with Recurrent Policy Gradients}.
\newblock In \emph{ICANN}, 2007.

\end{thebibliography}

\end{document}